\documentclass[conference]{IEEEtran}
\UseRawInputEncoding

\usepackage[utf8]{inputenc} 
\usepackage[T1]{fontenc}    
\usepackage{hyperref}       
\usepackage{url}            
\usepackage{booktabs}       
\usepackage{amsfonts}       
\usepackage{nicefrac}       
\usepackage{microtype}      
\usepackage{lipsum}
\usepackage{fancyhdr}       
\usepackage{graphicx}       
\usepackage{float}
\usepackage{subfig}
\usepackage{algpseudocode}
\usepackage{algorithm}
\usepackage{marvosym}

\newcommand{\BiStateTwo}[2]{%
  \State{\makebox[1.5cm]{#1\hfill}#2}%
  }
\newcommand{\BiState}[2]{%
\State{\makebox[2.5cm]{#1\hfill}#2}%
}

\begin{document}

\title{Predicting Future Mosquito Larval Habitats Using Time Series Climate Forecasting and Deep Learning}

\author{\IEEEauthorblockN{Christopher Sun}
\IEEEauthorblockA{\textit{Monta Vista High School}\\Cupertino, CA, United States\\chrisun1030@gmail.com}
\and
\IEEEauthorblockN{Jay Nimbalkar}
\IEEEauthorblockA{\textit{Green Hope High School}\\Cary, NC, United States\\jayomkarnimbalkar@gmail.com}
\and
\IEEEauthorblockN{Ravnoor Bedi}
\IEEEauthorblockA{\textit{Bronx High School of Science}\\New York City, NY, United States\\ravnoorbedi@gmail.com}}

\IEEEoverridecommandlockouts
\IEEEpubid{\makebox[\columnwidth]{978-1-6654-7345-3/22/\$31.00~\copyright2022 IEEE \hfill} \hspace{\columnsep}\makebox[\columnwidth]{ }}
\maketitle
\IEEEpubidadjcol

\begin{abstract}
Mosquito habitat ranges are projected to expand due to climate change. This investigation aims to identify future mosquito habitats by analyzing preferred ecological conditions of mosquito larvae. After assembling a data set with atmospheric records and larvae observations, a neural network is trained to predict larvae counts from ecological inputs. Time series forecasting is conducted on these variables and climate projections are passed into the initial deep learning model to generate location-specific larvae abundance predictions. The results support the notion of regional ecosystem-driven changes in mosquito spread, with high-elevation regions in particular experiencing an increase in susceptibility to mosquito infestation. 
\end{abstract}


\section{Introduction}
Mosquito habitats and breeding ranges are expanding globally \cite{colon}. Habitat preferences are based on the interaction of several factors, including temperature, humidity, rainfall, elevation, and availability of hosts. Climate change has been identified as a key driving factor for the shifts in mosquito distribution over the past 70 years and is likely to continue to be the chief determinant of mosquito population spread \cite{colon}. According to current trends, climate change will lead to major shifts in meteorological variables and land cover distributions, including an increase in average temperature, rising ocean levels, and increased severity of storms and droughts. 

Regional changes in climate have allowed for the expansion of mosquito populations to new environments. Mosquitoes bring detrimental vector-borne illnesses, such as the West-Nile virus and dengue. As warmer temperatures are correlated with accelerated mosquito development and illness spread, the necessity to predict potential outbreaks has become a priority \cite{NileVirusSpread}. 

Mosquito breeding sites require specific ecological conditions. Using habitat patterns such as rainwater pools, riverbed pools, streams, and marshes in conjunction with meteorological data allows for the assembly of larvae development models to forecast conditions that meet the habitable requirements \cite{MosquitoPatterns}. Using artificial intelligence, predictions can be scaled to adaptable national or global models to identify nuanced relationships between atmospheric variables and mosquito abundance \cite{mosquito control}. 

Previous research has not integrated these artificial intelligence techniques with region-specific analyses to forecast the particular ecosystems that will become most conducive to mosquito habitation. This investigation harnesses deep learning to develop a mosquito larvae abundance model and to conduct time series climate forecasting, with the goal of pinpointing the most likely locales and ecosystems of future mosquito infestations in the United States.
\subsection*{Research Overview}
The research described in this article was divided into three phases. The first phase involved gathering meteorological data and larvae counts from various locations in the United States and using this data set to create a predictive model for mosquito larvae abundance. The second phase involved extracting time series sequences of the said meteorological variables for specific regions of interest, to allow for the forecasting of environmental conditions. The third phase involved feeding these environmental predictions into the predictive model developed in the first phase to obtain quantitative measurements of mosquito larvae abundance and identify potential breeding spots for mosquitoes. 
\begin{figure}[H] 
    \centering
    \includegraphics[scale=0.36]{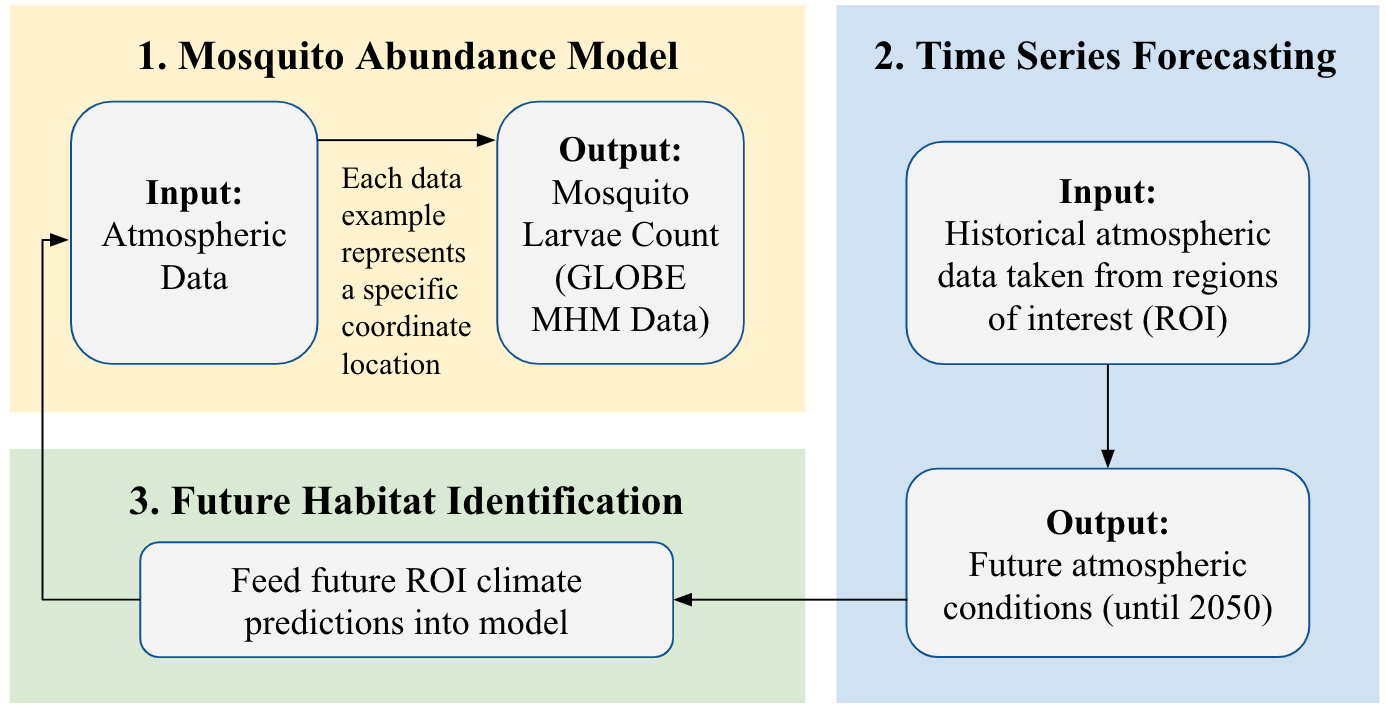}
    \caption{This flowchart details the sequential objectives of this work. A neural network is trained, after which climate forecasting creates future inputs that are passed into the neural network to generate mosquito larvae abundance predictions.}
\end{figure}

\section{Mosquito Larvae Abundance Model}
\subsection{Data Availability Statement}
Mosquito larvae abundance data was obtained from the GLOBE Mosquito Habitat Mapper (MHM) tool in the GLOBE Observer app, a worldwide citizen science app \cite{globe}. Meteorological data was obtained from the Weather Underground and Weather WX databases, commercial data providers with historical weather archives available on the web \cite{wunderground} \cite{weatherwx}. Historical time series sequences were obtained from the Climate Engine tool \cite{climate engine}. Particularly, the Gridded Surface Meteorological dataset (gridMET), derived from the PRISM satellite and the NLADS project, were used for temperature and precipitation records \cite{gridmet}. 

\subsection{Data Collection}
Mosquito observations are collected using the MHM tool, which records the type of water source in which larvae were observed. Possible categories include still water, flowing water, and water containers. Using a GLOBE data processing tool, mosquito larvae observations made from artificial containers (ovitraps) were filtered and removed to minimize the effect of opportunistic data on the true population densities of mosquito larvae \cite{streamlit}. This data cleaning process was grounded on the assumption that the result would be a valid data, as if gathered in the field for the purpose of analysis, rather than due to opportunistic circumstances. Larvae counts measured on the same dates in the same location were also treated as a singular observation. In each region, the monthly averages of daily maximum, mean, and minimum temperature, as well as precipitation amount and days of precipitation, were collected.\footnote{Historical ecological data was only available from airports. Mosquito larvae records from locations with no airports within a 30-mile vicinity were excluded.} Gathering monthly-averaged data was a reasonable decision due to the lag time that has been shown to exist between environmental phenomena and resulting trends in mosquito abundance \cite{lag}. The final data set was represented by 166 locations across the contiguous United States, containing the following numerical features: Average Daily Mean Temperature, Average Daily Maximum Temperature, Average Daily Minimum Temperature, Days of Precipitation, Average Daily Precipitation Amount, Elevation, and Larvae Count. 
\subsection{Model Architecture} \label{sec:mosquito model}
Due to the non-optimized nature of the GLOBE database, statistical transformations of mosquito larvae counts were necessary prior to predictive modeling. Log\textsubscript{10} transformation was applied to the mosquito larvae counts and other variables were standardized using z-scores.
\begin{figure}[H] 
    \centering
    \includegraphics[scale=0.27]{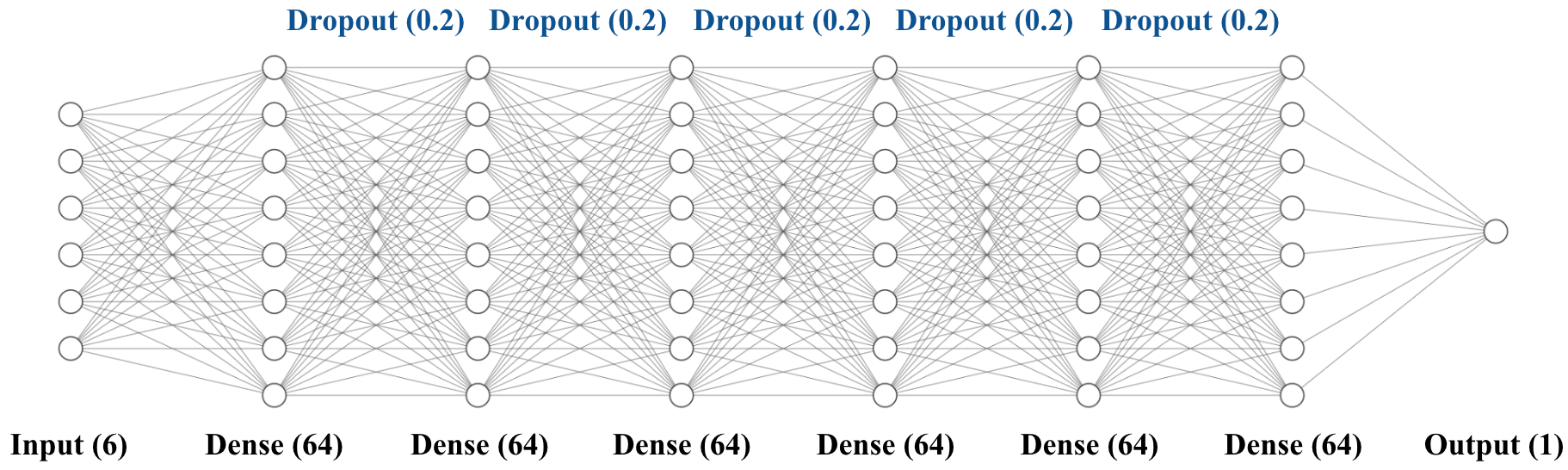}
    \caption{Shown is the architecture of the neural network used to predict mosquito larvae abundance. The model contained 21,313 parameters.}
\end{figure}
A deep neural network was assembled for the prediction of mosquito larvae counts. The model contained 6 dense layers with 64 hidden nodes each, followed by an output node. The rectified linear activation unit was used between dense layers. Additionally, dropout layers with a dropout probability of 0.2 were applied between dense layers to regularize the model. The model was trained until convergence using Adam optimization, Xavier weight initialization, and a mini-batch size of 8 examples.

Thirty-five of the oldest data examples were withheld as validation data to gauge whether the model could backcast previous mosquito larvae counts from historical ecological data. 
\subsection{Model Performance}
\begin{figure}[H] 
    \centering
    \hspace*{-7mm}\includegraphics[scale=0.34]{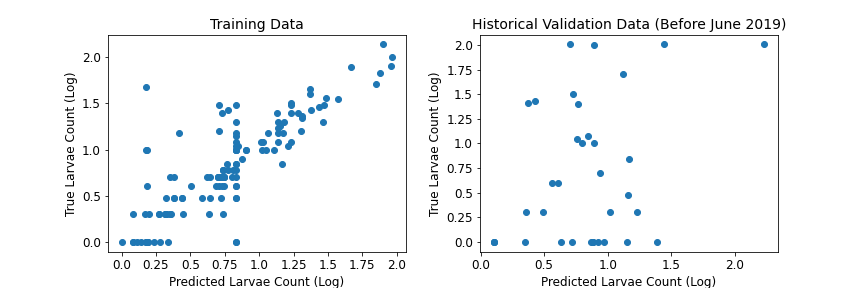}
    \caption{The neural network converged on training data but achieved a moderate positive correlation on historical validation data. The degree of generalization could be improved with cleaner and larger data sources.}
\end{figure}
\begin{table}[H]
    \centering
    \renewcommand{\arraystretch}{1.3}
    \caption{Model Metrics: Correlations and Their Statistical Significance}
    \begin{tabular}{|l|l|l|}
    \hline
                        & \textbf{Training} & \textbf{Validation} \\ \hline\hline
    \textbf{R}          & 0.888             & 0.489               \\ \hline
    \textit{\textbf{P}} & 1.26E-45         & 1.44E-3             \\ \hline
    \end{tabular}
\end{table}
The correlation coefficient between training data labels and model predictions was 0.888. However, the correlation coefficient between validation data and model predictions was 0.488. In other words, the deep neural network was able to understand the intricacies of the training data but fell short when it came to generalizing to unseen data. Though a moderate positive correlation existed between predicted larvae counts and ground-truth larvae counts on validation data, there were instances of large residuals between these values. In practice, positive residuals are more concerning than negative residuals, since a negative residual corresponds to an overestimate of mosquito larvae abundance. This may lead to extra mosquito-prevention precautions being taken when not necessary, which is not directly harmful except from a resource-conservation standpoint. On the other hand, a positive residual implies an underestimate of mosquito larvae abundance, which could lead to a false perception of safety. This may lead to vector-borne disease breakouts due to necessary precautions not being instituted. The predictions of the deep neural network reveal more negative than positive residuals, meaning the model tended to liberally flag locations as containing high mosquito larvae abundances, when in truth they were of less concern. 

\section{Climate Forecasting}
\subsection{Time Series Analysis}
Temperature and precipitation data for the summer months between 1979 and 2021 were acquired using the Climate Engine database for all 48 contiguous states. Measurements were averaged spatiotemporally across the 3-month summertime period from June 22 to September 22 for each state.

Prior to using a Long Short-Term Memory network (LSTM) for time series forecasting, it was discovered that the trends in temperature and precipitation somewhat conformed to a pattern resembling the following periodic function, where $T$ is the target atmospheric variable given the year $t$ since the initial year $t_0$.
\begin{equation}
T(t) = \lambda t - e^{-\alpha t}\sin(\theta t)\gamma t^\beta + \phi.
\end{equation}
The approximate parameters that can be used to estimate the trends are as follows:
$$\lambda\approx0.01, \alpha\approx-0.01, \theta\approx0.6, \gamma\approx0.5, \beta\approx0.03, \phi\approx T(t_0).$$ 
This finding confirmed the validity of a supervised deep learning approach to conduct time-series forecasting.

It was also discovered that minimum and maximum temperature shared a high correlation with mean temperature. In particular, 
\begin{equation}\label{eq:2}
T_{min}(t) = T_{mean}(t) - k_{min}, \forall t \in \{t_i\}_{i=0}^n
\end{equation}
\begin{equation}\label{eq:3}
T_{max}(t) = T_{mean}(t) + k_{max}, \forall t \in \{t_i\}_{i=0}^n,
\end{equation}
where $k_{min}$ is a constant of adjustment between the minimum and mean temperature and $k_{max}$ is a constant of adjustment between the maximum and mean temperature. Hence, the climate forecasting task for minimum and maximum temperatures was simplified into the following problem: Find $k$ such that $MAE = \frac{1}{n}\sum_{i=0}^{n}|T(t_i) - S_i|$ is minimized, where $T(t_i)$ is the temperature predicted using the corresponding function above and $S_i$ is the true temperature.
\subsection{LSTM Networks}
\begin{figure}[H]
    \centering
    \includegraphics[scale=0.35]{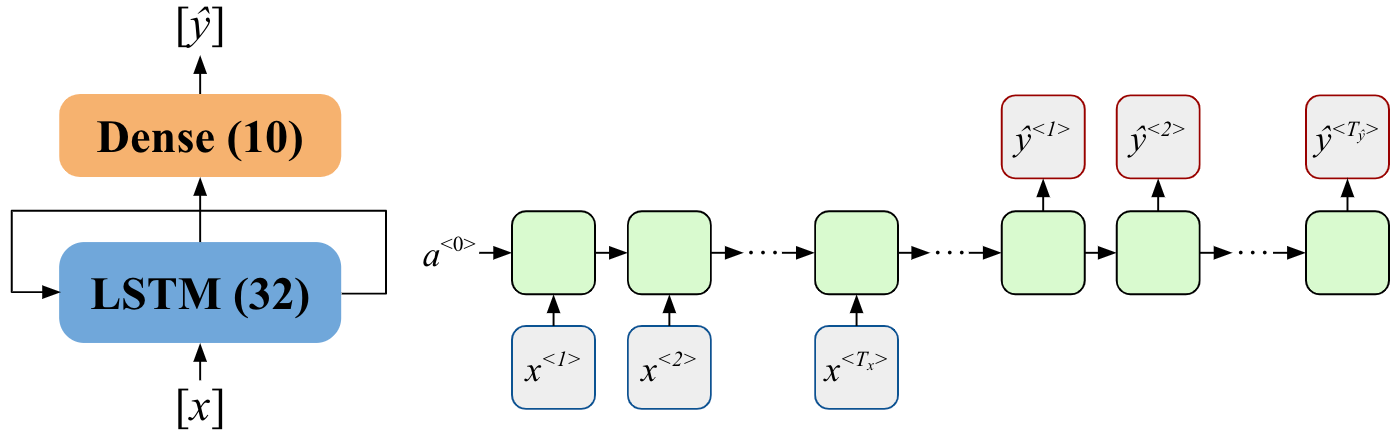}
    \caption{Shown are the architecture of the LSTM network (left) and an illustration of process of time-stepping (right). For our application, $T_x$ equaled 20 and $T_{\hat{y}}$ equaled 10. The model contained 4,682 parameters.}
\end{figure}

An LSTM network with 32 LSTM units followed by a dense layer was trained on the aforementioned temperature and precipitation sequences to conduct climate forecasting. The model's input length was 20 time steps (corresponding to the past 20 years) and the output length was 10 time steps (corresponding to the next 10 years). Dropout regularization with a dropout probability of 0.2 was applied before the LSTM layer to prevent overfitting. The model was trained until convergence using the same specifications as described in Section \ref{sec:mosquito model}.

The following algorithm details the process used to generate future climate predictions, where $m$ is the number of locations, $l$ is the length of the lookback sequence, $p$ is the length of the prediction sequence, $t$ is the number of future sequences (each of length $p$), $f: x \rightarrow y$ is a trained LSTM model, $X$ is the array $\{[X_{11} \dots X_{1l}], [X_{21} \dots X_{2l}], \dots, [X_{m1} \dots X_{ml}]\}$, and $Y$ is the output array.
\begin{algorithm}[H]
    \caption{Climate Forecasting Process}
    \begin{algorithmic}
    \Require{$X$} \Comment{sequence of $m$ arrays of length $l$}
    \Ensure{$Y$} \Comment{prediction sequence of $m$ arrays of length $p \cdot t$}
    \Function{Forecast}{X}
        \For{$i \gets 0$ to $m$}
            \State $x \gets X_i$
            \State $Y_i \gets [\:]$
            \BiStateTwo{$\mu \gets \mu_{x}$}{$\sigma \gets \sigma_{x}$}
            \State$x \gets \frac{x-\mu}{\sigma}$
            \For{$j \gets 0$ to $t$} 
                \State $y \gets f(x)$
                \BiState{$y \gets y \cdot \sigma + \mu$}{$x \gets x \cdot \sigma + \mu$}
                \State $Y_i \gets Y_i\,||\,y$
                \State $x \gets x[(p-l):\:]\,||\,y$
                \BiStateTwo{$\mu \gets \mu_{x}$}{$\sigma \gets \sigma_{x}$}
                \State $x \gets \frac{x-\mu}{\sigma}$
            \EndFor
            \State $Y \gets Y||Y_i$
        \EndFor
        \State \Return {Y}
    \EndFunction
    \end{algorithmic}
\end{algorithm}
\vspace{2mm}
Average summer temperature and average summer precipitation levels were forecasted until the year 2050. Then, maximum and minimum monthly temperatures were derived from average monthly temperatures using equations \ref{eq:2} and \ref{eq:3}. The monthly days of precipitation for a region could not be forecasted due to the absence of available data on this variable. As a result, due to a strong positive correlation between monthly days of precipitation and average precipitation amount, the former was predicted from the latter using a linear model. Elevation changes were not forecasted; future inputs were simply current mean elevations of each region. These six variables were then passed into the neural network described in Section \ref{sec:mosquito model} to predict future mosquito larvae abundance. 

\section{Future Mosquito Habitat Identification}
\subsection{United States Mosquito Larvae Abundance Forecasts}
\begin{figure}[H] 
    \centering
    \includegraphics[scale=0.45]{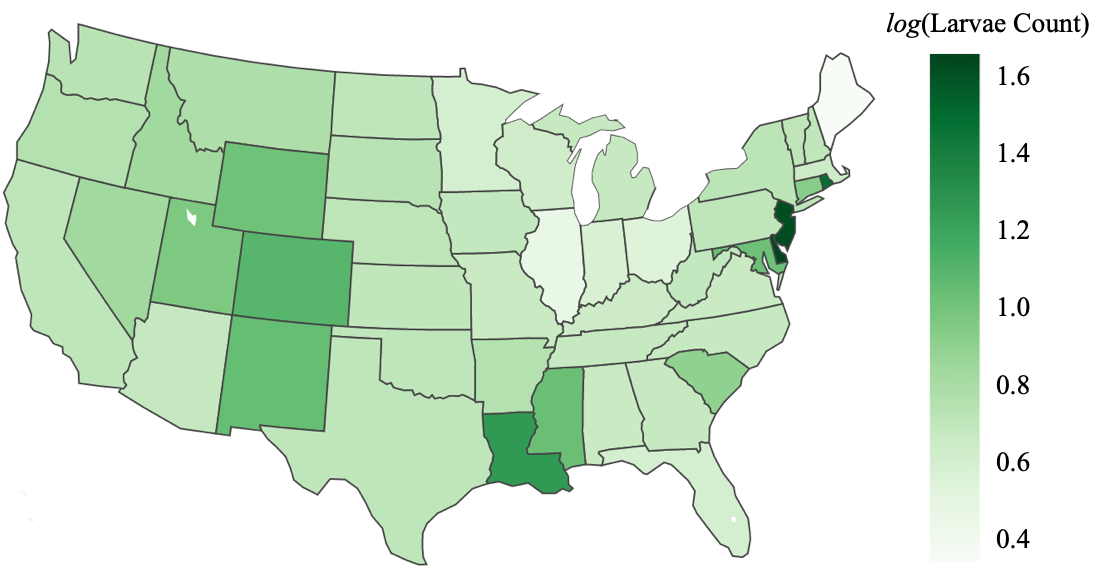}
    \caption{Larvae Abundance by State in 2050. This map displays the model's projections for larvae abundance colored on a logarithmic scale.}
\end{figure}
After conducting climate forecasting for each state, mosquito larvae abundance projections were obtained and displayed in the choropleth map in Fig. 5. Our results show that by 2050, the Rocky Mountain states will emerge as likely locations of mosquito breeding. Namely, Colorado, Utah, Wyoming, and New Mexico will contain higher larvae counts than neighboring states as well as the US average larvae count. This result is interesting considering the said states have high elevations, which is an attribute assumed to have low correlations with mosquito abundance. However, upon closer inspection, these projections align with previous research findings, such as those of Derek-Scasta (2021), who discovered that changing weather variability could shift some mosquito species into higher elevations, causing mosquito densities at high-altitude regions to exceed densities at mid-altitude regions \cite{high elevation}. In addition, states along the Atlantic coastline are projected to contain varying mosquito larvae abundances, but results from Louisiana and New Jersey contrast with results from Florida and Maine, thus preventing the generalization of a relationship between coastlines and mosquito larvae abundance.
\subsection{Region of Interest: Texas}
Due to the observed correlation between elevation and larvae abundance, the state of Texas was selected as a region of interest for further analysis at the regional level. Texas was chosen since it contains a steep longitudinal elevation gradient ranging from the Rocky Mountains to the Gulf of Mexico coast. Climate forecasting was conducted for each of Texas's ten CONUS climate divisions \cite{CONUS}. Segmentation by climate division permitted a more accurate regional characterization of Texas's diverse ecology, which allowed any observed shifts in larvae counts to be better attributed to a particular ecosystem. 
\begin{figure}[H] 
    \centering
    \includegraphics[scale=0.4]{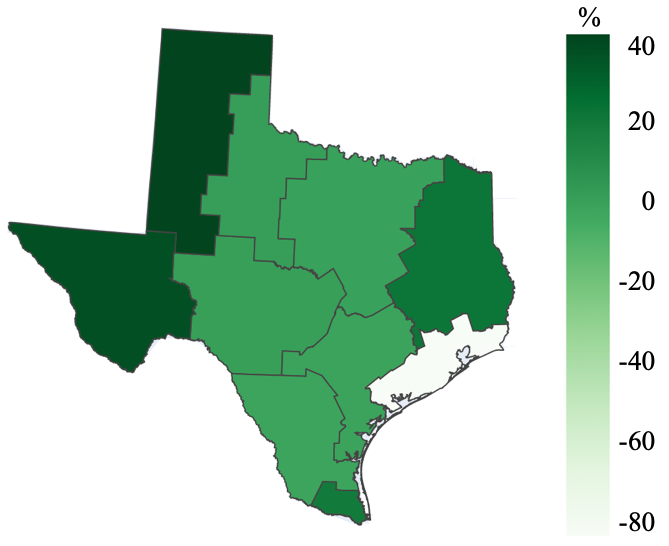}
    \caption{Percent Change in Larvae Abundance in Texas Climate Divisions (2030-2050).}
\end{figure}
Results for Texas confirmed the patterns observed at the national level. Namely, the two climate divisions with the highest elevation, Trans Pecos and High Plains, were projected to experience the greatest rate of increase in larvae abundance. To understand how ecological variables were correlated with this observation, temperature and precipitation were also mapped, as shown in Fig. 7.
\begin{figure}[H] 
    \centering
    \includegraphics[scale=0.38]{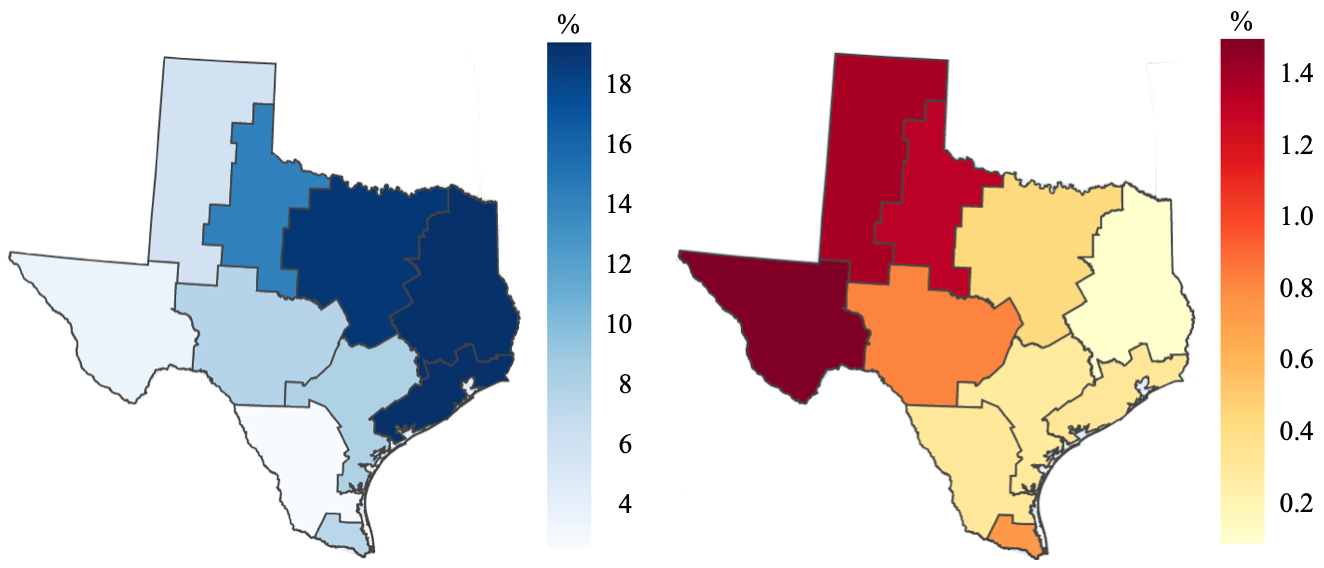}
    \caption{Meteorological Changes in Texas Climate Divisions (2030-2050). These maps show climate forecasts for precipitation (left) and temperature (right) as a percent change.}
\end{figure}
Between 2030 and 2050, precipitation is projected to increase more so in the eastern portion of Texas compared to the southern and western portions. The opposite trend was observed for mean temperature. In fact, western regions were projected to experience the highest rate of temperature increase. The likely relationship between these two factors and larvae abundance is that warmer temperatures in western Texas will intensify and prolong drought conditions, stalling precipitation over the next few decades, while making high-altitude conditions more favorable for the habitation of mosquitoes. 
\subsection{Analysis of Error}
The mosquito larvae abundance model's inability to generalize optimally may have been caused by error in the original GLOBE Mosquito Habitat Mapper data set. Locations with high larvae counts were outliers by nature, causing the model's threshold between underfitting and overfitting to be very steep, hence the suboptimal degree of generalization to validation data. Out of 35 validation data examples, the examples that were underestimates of mosquito larvae abundance mostly corresponded to locations where larvae abundance data was scarce. Hence, the quality of the model could be improved by simply augmenting the size of the data set, which is feasible in the near future due to its nature as a citizen science project.

\section{Conclusions}
This article aims to predict the abundance of mosquito larvae across the United States in the year 2050. To achieve this purpose, a data set consisting of citizen-collected mosquito larvae counts and several accompanying atmospheric and spatiotemporal variables is compiled. Then, atmospheric variables are analyzed to identify the conditions most suited to the habitation of mosquito larvae using a deep learning framework. Next, these variables are forecasted using an LSTM model to project future climatic conditions. Finally, these atmospheric projections are inputted back into the original deep learning model to obtain the desired mosquito larvae abundance predictions.

The results from this experiment support the idea that mosquito spread is largely location and ecosystem-dependent, which points to the benefits of utilizing localized citizen-science observations and conducting regional examinations. One finding was that states along the Rocky Mountain Range, which contain some of the highest elevations around the country, were predicted to have the highest larvae abundance in 2050. This observation was further supported by the case study of Texas, which predicted the greatest change in larvae counts to occur in the high-altitude western region. These results show that the greatest shifts in mosquito larvae abundances will occur in high-altitude locales. The most likely cause of this trend is that an increase in temperatures will render high-altitude regions warm enough for mosquito habitation for the first time. Precipitation and proximity to large bodies of water, however, did not appear to have a generalized correlation with larvae abundance, yielding varying larvae counts across the US. It is likely that unusually large larvae count observations in regions along the coastline were due to the presence of lurking variables such as high population densities rather than due to the coastal location itself. 

These findings point to the need for increased resource allocation to high-elevation areas to contain mosquito spread and resultant vector-borne diseases, since these locations are forecasted to become high-risk targets in the future. In addition, because there is a meager presence of mosquitoes in high-altitude regions today, the awareness and containment protocols in these areas regarding mosquitoes are likely lacking, which may lead to grave future consequences if no action is taken. 
\section*{Acknowledgment}
The authors would like to acknowledge the support of the 2022 Earth Explorers Team, NASA STEM Enhancement in the Earth Sciences (SEES) Virtual High School Internship program, and Dr. Russanne Low. The NASA Earth Science Education Collaborative (NESEC) leads Earth Explorers through an award to the Institute for Global Environmental Strategies, Arlington, VA (NASA Award NNX6AE28A). The SEES High School Summer Intern Program is led by the Texas Space Grant Consortium at the University of Texas at Austin (NASA Award NNX16AB89A). 
\section*{Additional Information}
CS, JN, and RB compiled data, performed research methodologies, and prepared this article. The authors have made the code used in this research available on \href{https://github.com/csun365/Mosquito-Habitat-Prediction}{GitHub (click)} \cite{code}. Data will be made available upon request.


\begin{thebibliography}{1}
\bibitem{colon}
F. J. Colón-González et al., ``Projecting the risk of mosquito-borne diseases in a warmer and more populated world: a multi-model, multi-scenario intercomparison modelling study,'' The Lancet Planetary Health, 5(7), e404-e414, July 2021.

\bibitem{NileVirusSpread}
Environmental Protection Agency. EPA. www.epa.gov/climate-indicators/climate-change-indicators-west-nile-virus 

\bibitem{MosquitoPatterns}
F. P. Amerasinghe and N. G. Indrajith, ``Postirrigation breeding patterns of surface water mosquitoes in the Mahaweli Project, Sri Lanka, and comparisons with preceding developmental phases,'' Journal of medical entomology, 31(4), 516-523, July 1994.

\bibitem{mosquito control}
A. Joshi and C. Miller, ``Review of machine learning techniques for mosquito control in urban environments,'' Ecological Informatics, 61, 101241, March 2021.

\bibitem{globe}
R. Low, ``Citizen Scientists as Community Agents of Change: GLOBE Observer Mosquito Habitat Mapper,'' In AGU Fall Meeting Abstracts (Vol. 2018, pp. IN22B-03).

\bibitem{wunderground}
Local weather forecast, news and conditions. Weather Underground. www.wunderground.com/

\bibitem{weatherwx}
WeatherWX.com, www.weatherwx.com/ 

\bibitem{climate engine}
J. L. Huntington et al., ``Climate engine: Cloud computing and visualization of climate and remote sensing data for advanced natural resource monitoring and process understanding,'' Bulletin of the American Meteorological Society, 98(11), 2397-2410, November 2017.

\bibitem{gridmet}
J. T. Abatzoglou, ``Development of gridded surface meteorological data for ecological applications and modelling,'' International Journal of Climatology, 33(1), 121-131, January 2013.

\bibitem{streamlit}
M. Kimura, ``Globe Data Dashboard,'' 2021. piphi5-globe-data-dashboard-srcmain-8lo71r.streamlitapp.com

\bibitem{lag}
K. Chang et al., ``Time-lagging interplay effect and excess risk of meteorological/mosquito parameters and petrochemical gas explosion on dengue incidence,'' Scientific reports, 6(1), 1-10, October 2016.

\bibitem{high elevation}
J. Derek Scasta, ``Livestock parasite management on high-elevation rangelands: ecological interactions of climate, habitat, and wildlife,'' Journal of Integrated Pest Management, 6(1), March 2015.

\bibitem{CONUS}
CONUS Climate Divisions. National Centers for Environmental Information, NOAA. www.ncei.noaa.gov/access/monitoring/reference-maps/conus-climate-divisions

\bibitem{code}
C. Sun, J. Nimbalkar, and R. Bedi, ``Github repository for the research paper 'Predicting Future Mosquito Larval Habitats Using Time Series Climate Forecasting and Deep Learning,''' github.com/csun365/Mosquito-Habitat-Prediction
\end{thebibliography}
\end{document}